\pdfoutput=1
\PassOptionsToPackage{table}{xcolor}
\documentclass[11pt]{article}

\usepackage[final]{acl}

\usepackage{times}
\usepackage{latexsym}
\usepackage{multirow}
\usepackage{amssymb}
\usepackage{subcaption}

\usepackage{amsmath}

\usepackage[T1]{fontenc}
\usepackage[utf8]{inputenc}
\usepackage{microtype}
\usepackage{inconsolata}
\usepackage{tabularx}             
\usepackage{booktabs}             

\usepackage{graphicx}
\usepackage[table]{xcolor}
\usepackage{booktabs}
\usepackage{tcolorbox}

\title{GraphLoRA: Structure-Aware Low-Rank Adaptation for Large Language Model Recommendation}

\author{
  \textbf{Lin Mu\textsuperscript{1}},
  \textbf{Guoji Wang\textsuperscript{1}},
  \textbf{Li Ni\textsuperscript{1}},
  \textbf{Lei Sang\textsuperscript{1}},
  \textbf{Zhize Wu\textsuperscript{2}},
  \\
  \textbf{Peiquan Jin\textsuperscript{3}},
  \textbf{Yiwen Zhang\textsuperscript{1}\footnotemark[1]}
  \\
  \textsuperscript{1}Anhui University,
  \textsuperscript{2}Hefei University,
  \\
  \textsuperscript{3}University of Science and Technology of China,
\\
  \small{
    \texttt{\{mulin, nili, sanglei, zhangyiwen\}@ahu.edu.cn \{wangguoji\}@stu.ahu.edu.cn}
  }
  \\
  \small{
    \texttt{wuzz@hfuu.edu.cn jpq@ustc.edu.cn}
  }
}

\begin{document}
\maketitle

\begin{abstract}
Large Language Models (LLMs) have shown strong potential for recommendation (LLMRec) due to their powerful reasoning and generalization abilities. However, effectively aligning the textual semantics modeled by LLMs with the collaborative signals remains a key challenge. Existing methods either translate collaborative information into textual prompts or inject pre-trained embeddings into the LLM, both of which treat structural information as static input and fail to capture high-order relational dependencies.
To bridge this gap, we propose \textbf{GraphLoRA}, a novel framework that generalizes low-rank adaptation from independent to structure-aware propagation. GraphLoRA embeds a trainable graph message-passing network within the low-rank adaptation pathway, enabling structural signals to propagate through the parameter space.
This design allows collaborative topology to explicitly guide parameter updates, fostering deep integration between graph-structured and textual semantic information. Extensive experiments on multiple benchmarks demonstrate that GraphLoRA not only outperforms state-of-the-art LLM-based recommendation methods but also achieves superior generalization, effectively balancing structural reasoning capability with computational efficiency. 
Code is available at \href{https://github.com/wgj15965/GraphLoRA}{https://github.com/wgj15965/GraphLoRA}. 
\end{abstract}

\renewcommand{\thefootnote}{\fnsymbol{footnote}}
\footnotetext[1]{Corresponding author}

\begin{figure}[t] 
  \centering
  \begin{subfigure}[b]{\linewidth} 
    \centering
    \includegraphics[width=0.8\linewidth]{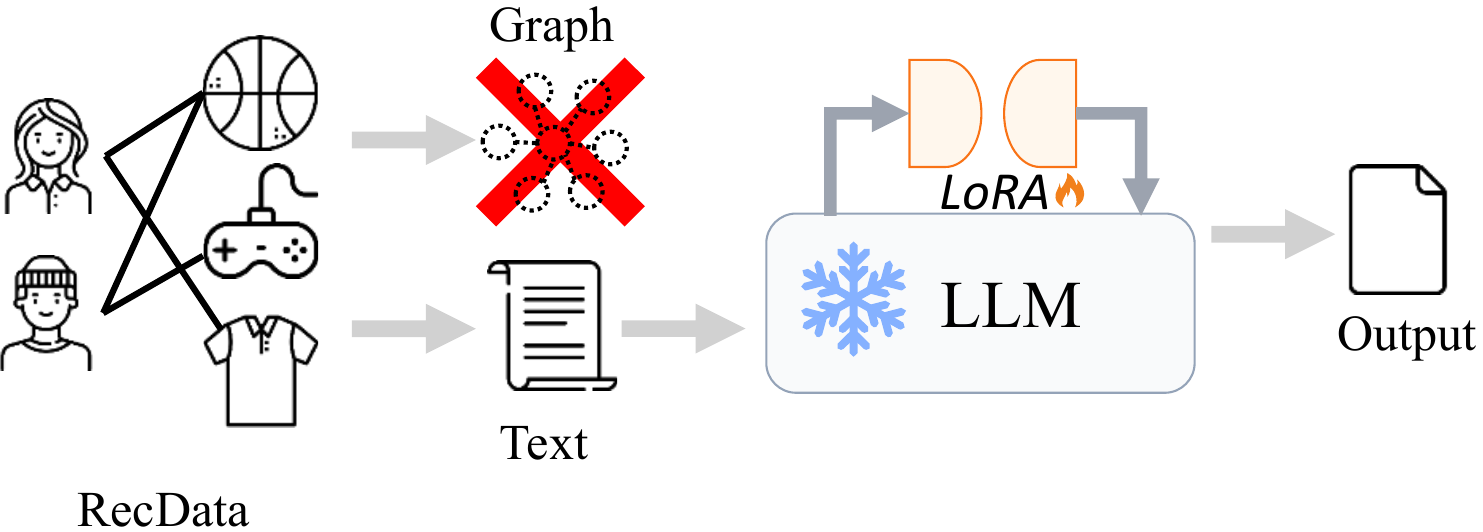} 
    \caption{Input-Space Alignment (e.g., TALLRec)}
    \label{fig:input_align}
  \end{subfigure}
  \vspace{0.3cm} 
  \begin{subfigure}[b]{\linewidth}
    \centering
    \includegraphics[width=0.8\linewidth]{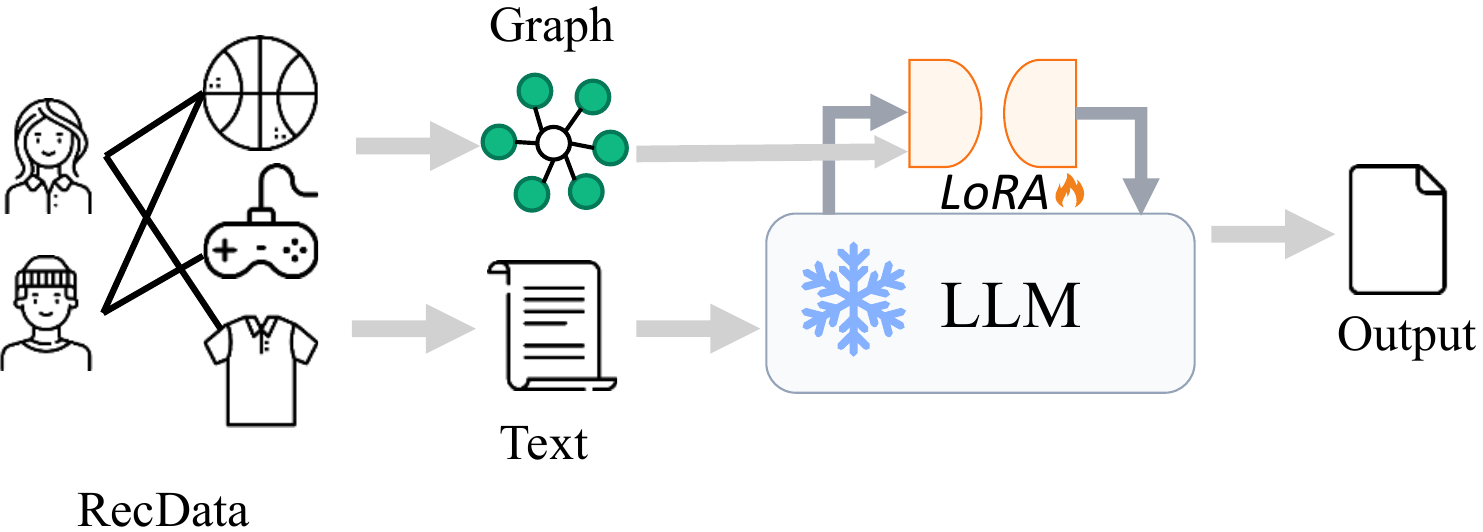}
    \caption{Parameter-Space Alignment (e.g., CoRA)}
    \label{fig:param_align}
  \end{subfigure}
  \vspace{0.3cm} 
  \begin{subfigure}[b]{\linewidth}
    \centering
    \includegraphics[width=0.8\linewidth]{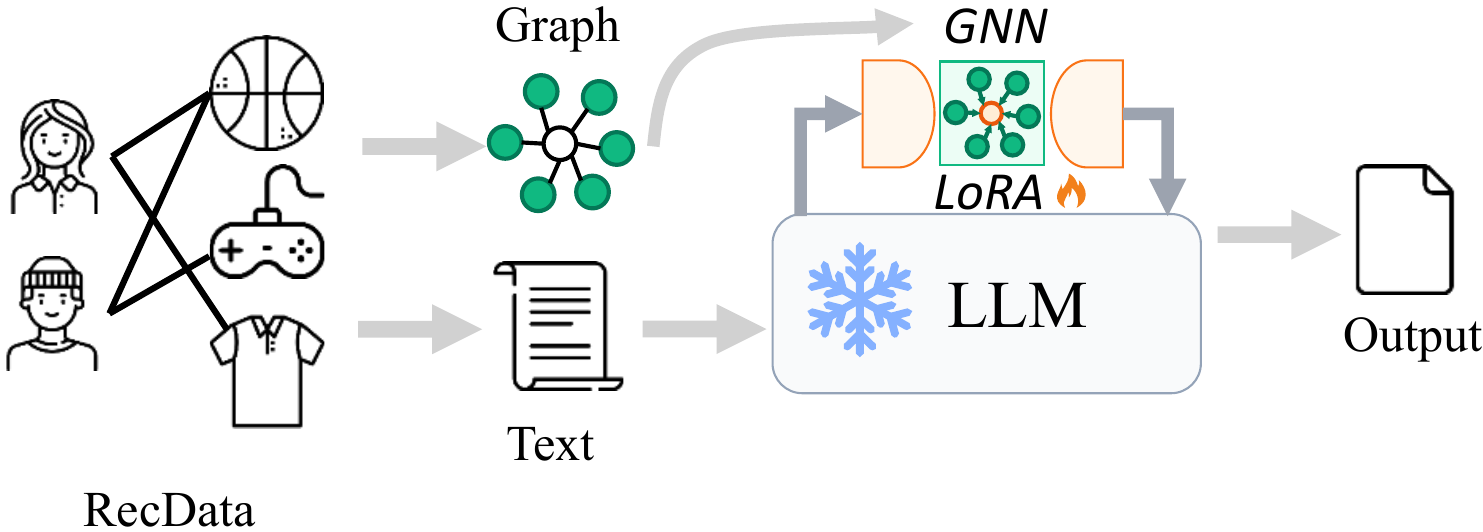}
    \caption{Structure-Aware Alignment (Ours)}
    \label{fig:struct_align}
  \end{subfigure}
  \caption{Comparison of collaborative alignment paradigms. 
\textbf{(a) Input-Space Alignment} converts interaction histories into textual prompts for the LLM input.
\textbf{(b) Parameter-Space Alignment} injects static, externally encoded embeddings into LLM weights (e.g., via LoRA).
\textbf{(c) Structure-Aware Alignment (Ours)} integrates a learnable GNN within the LoRA bottleneck (between $A$ and $B$) to perform dynamic message passing directly in the parameter space.}
  \label{fig:motivation}
\end{figure}

\section{Introduction}
Recommender systems play a central role in personalized information access, enabling users to efficiently discover relevant content~\cite{wu2022graph}. Recently, large language models (LLMs) have emerged as powerful recommenders due to their remarkable capabilities in reasoning, generalization, and contextual understanding~\cite{liu2025llmemb, wu2024survey}. However, a fundamental challenge persists: effectively align textual semantics—naturally handled by LLMs—with collaborative information derived from user–item interactions, which is inherently structured. Bridging this gap is essential for unleashing the full potential of LLM-based recommendation.

Early efforts in LLM-based recommendation primarily relied on \textit{Input-Space Alignment} (see Figure~\ref{fig:motivation}(a)), where collaborative information was translated into prompts or input tokens. For example, TALLRec~\cite{TALLRec} linearized user-item interaction histories into textual prompts, while CoLLM~\cite{Collm2025} projected user and item embeddings-obtained from matrix factorization (MF)-into soft tokens. Although these methods allow LLMs to access collaborative signals, they force the model to passively “read” structural information rather than internally reason about relational dependencies. Moreover, linearized or tokenized representations struggle to capture the complex, high-order topology of user–item interaction graphs, leading to a loss of structural inductive bias~\cite{huang2024survey}.

To strengthen internal modeling, recent research has pivoted towards \textit{parameter-space alignment}, as illustrated in Figure~\ref{fig:motivation}(b) (e.g., CoRA~\cite{CoRA}). Instead of modifying textual input, CoRA directly injects collaborative signals into LLM's parameters (e.g., via LoRA), enabling the model to internalize user–item relationships within its parameter space. This design overcomes the limitations of input-space alignment by allowing the LLM to encode collaborative patterns rather than merely read them. However, CoRA still depends on externally pre-trained embeddings (e.g., from matrix factorization) that are injected as static weights. Consequently, structural information remains external and cannot be jointly refined with the model’s semantic representations. To address this limitation, we advocate for \textit{structure-aware alignment} within the LLM itself. We reinterpret LoRA not as a passive fine-tuning module, but as a learnable reasoning pathway capable of propagating collaborative signals in the parameter space. This perspective generalizes the conventional LoRA formulation—from independent low-rank adaptation to structure-aware low-rank propagation—thereby enabling the LLM to internalize collaborative topology and jointly refine semantic and structural representations.

Building upon this insight, we propose \textbf{GraphLoRA}, a novel framework that realizes this \textit{\textbf{structure-aware alignment}} (Figure~\ref{fig:motivation}(c)).
GraphLoRA embeds a differentiable graph message-passing module within the LoRA bottleneck (between $A$ and $B$), constructing a learnable structural pathway directly in the low-rank latent space. Specifically, collaborative representations are first projected into the low-rank latent space, where high-order neighborhood information is dynamically aggregated through graph message passing, and the enriched structural embeddings are then projected back to the parameter space to update the model.
This mechanism allows collaborative information to explicitly guide parameter updates, fostering deep interaction between collaborative topology and semantic understanding.
Through joint optimization of the structural encoder and the LLM adaptation under a unified objective, GraphLoRA effectively distills neighborhood-aware structural signals—achieving deep structure–semantic synergy while preserving the LLM’s linguistic competence.

The contributions can be summarized as follows:
\begin{itemize}
    \item We propose \textbf{GraphLoRA}, which integrates a graph message-passing module within the LoRA bottleneck, generalizing independent adaptation to structure-aware propagation and enabling LLMs to internalize collaborative topology.
    
    \item We design a joint optimization framework that couples graph-structured reasoning with semantic adaptation, yielding representations that are both structure-aware and linguistically coherent.
    
    \item Extensive experiments on multiple benchmark datasets demonstrate that GraphLoRA consistently outperforms state-of-the-art LLM-based recommendation methods under compact parameter settings. Additional ablation and efficiency studies further confirm its robustness, parameter efficiency, and structural interpretability across diverse recommendation scenarios.
\end{itemize}

\section{Related Works}

\subsection{LLM-based Recommendation}
The application of large language models (LLMs) in recommender systems has evolved rapidly. Early works primarily utilized in-context learning~\cite{incontextlearning}, leveraging the extensive world knowledge of LLMs via carefully crafted prompts to perform recommendation tasks without parameter updates~\cite{gao2023chat, dai2023uncovering}. While cost-effective, these methods often struggle to capture domain-specific collaborative signals.

To address this, subsequent research shifted towards instruction tuning with \textit{input-space alignment}. 
Foundational frameworks like P5~\cite{geng2022recommendation} laid the groundwork by unifying recommendation tasks into sequence generation. Following this, a representative method, TALLRec~\cite{TALLRec}, reformulated interaction data into linear textual instructions. To further incorporate dense collaborative signals, methods like CoLLM~\cite{Collm2025} and LlaRA~\cite{liao2024llara} embedded user/item ID embeddings from external encoders directly into the input prompt as soft tokens, while BinLLM~\cite{zhang2024binllm} encoded such information into binary sequences. However, this paradigm often results in a separation between collaborative modeling and LLM reasoning. Mapping complex graph topology into linear sequences or static tokens can impose an information bottleneck, effectively flattening the structural context before it reaches the LLM.

\subsection{Graph-Enhanced LLMs and Structure-Aware Tuning}
Recognizing the importance of structural data, researchers have explored integrating graph neural networks (GNNs)~\cite{kipf2017semi} with LLMs. One line of research adopts a projector-based approach, where a graph encoder is aligned with the LLM via a projection layer. For instance, GraphGPT~\cite{tang2024graphgpt} and HiGPT~\cite{tang2024higpt} employed graph instruction tuning to align graph structural knowledge with the LLM's token space, while GIMLET~\cite{zhao2023gimlet} unified modalities for molecule tasks. Despite their effectiveness, tuning the projector or the entire model can be computationally expensive.

Another emerging direction focuses on parameter-space alignment via parameter-efficient fine-tuning (PEFT)~\cite{ding2023parameter}. Although general-purpose PEFT methods like Adapter~\cite{houlsby2019parameter} and Prefix-Tuning~\cite{li2021prefix} have proven effective in NLP, and recent advancements continue to optimize the density and efficiency of adaptation matrices (e.g., DenseLoRA~\cite{denselora}), recommendation works specifically leverage low-rank adaptation (LoRA)~\cite{hu2022lora} to modify model weights. A representative method, CoRA~\cite{CoRA}, injected collaborative signals directly into these adaptation matrices.
To bridge this gap, our proposed GraphLoRA embeds a trainable message-passing GNN within the LoRA bottleneck, thereby enabling dynamic aggregation of high-order structural signals directly in the parameter space.

\section{Background}

\subsection{Low-Rank Adaptation (LoRA)}
LoRA~\cite{hu2022lora} adapt LLMs by freezing pre-trained weights while injecting trainable rank decomposition matrices. For a linear layer with pre-trained weights $\mathbf{W}_0 \in \mathbb{R}^{d_{model} \times d_{model}}$, the weight update is parameterized by two low-rank matrices: a down-projection $\mathbf{A} \in \mathbb{R}^{r \times d_{model}}$ and an up-projection $\mathbf{B} \in \mathbb{R}^{d_{model} \times r}$, where $r \ll d_{model}$ is the intrinsic rank.
The forward pass for an input hidden state $\mathbf{x} \in \mathbb{R}^{d_{model}}$ is computed as:
\begin{equation}
    \mathbf{h} = \mathbf{W}_0 \mathbf{x} + \frac{\alpha}{r} \mathbf{B} \mathbf{A} \mathbf{x},
\end{equation}
where $\alpha$ is a scaling scalar. In GraphLoRA, we intervene in the latent space between $\mathbf{A}$ and $\mathbf{B}$ to inject collaborative signals.

\subsection{Message-Passing GNNs} \label{sec:mpg}
Recommendation data typically forms a user-item bipartite graph $\mathcal{G}=(\mathcal{U}, \mathcal{I}, \mathcal{E})$. To capture high-order collaborative signals without conflicting with LoRA's notation, we describe GNNs using the general message passing neural network (MPNN) paradigm~\cite{gilmer2017neural}.
MPNN defines structural learning via local aggregation. For a node $u$, its representation $\mathbf{e}_u^{(l)}$ at layer $l$ is updated by aggregating messages from its neighbors $\mathcal{N}_u$:
\begin{equation}
    \mathbf{m}_u^{(l)} = \text{AGG}\left(\left\{ \phi(\mathbf{e}_v^{(l)}, \mathbf{e}_u^{(l)}) : v \in \mathcal{N}_u \right\}\right),
\end{equation}
\begin{equation}
    \mathbf{e}_u^{(l+1)} = \psi \left(\mathbf{e}_u^{(l)}, \mathbf{m}_u^{(l)}\right),
\end{equation}
where $\phi(\cdot)$ is the message function, $\text{AGG}(\cdot)$ is a permutation-invariant aggregation operator (e.g., Mean, Sum), and $\psi(\cdot)$ is the update function. This unified view encompasses both general graph encoders (e.g., GraphSAGE~\cite{hamilton2017inductive}, GAT~\cite{velickovic2018graph}) and recommendation-specific variants (e.g., LightGCN~\cite{he2020lightgcn}, NGCF~\cite{wang2019neural}). 
In this work, we adopt the latter category as the structural backbone, given their proven effectiveness in modeling collaborative filtering signals.

\begin{figure*}[t]
  \centering
  \includegraphics[width=0.9\linewidth]{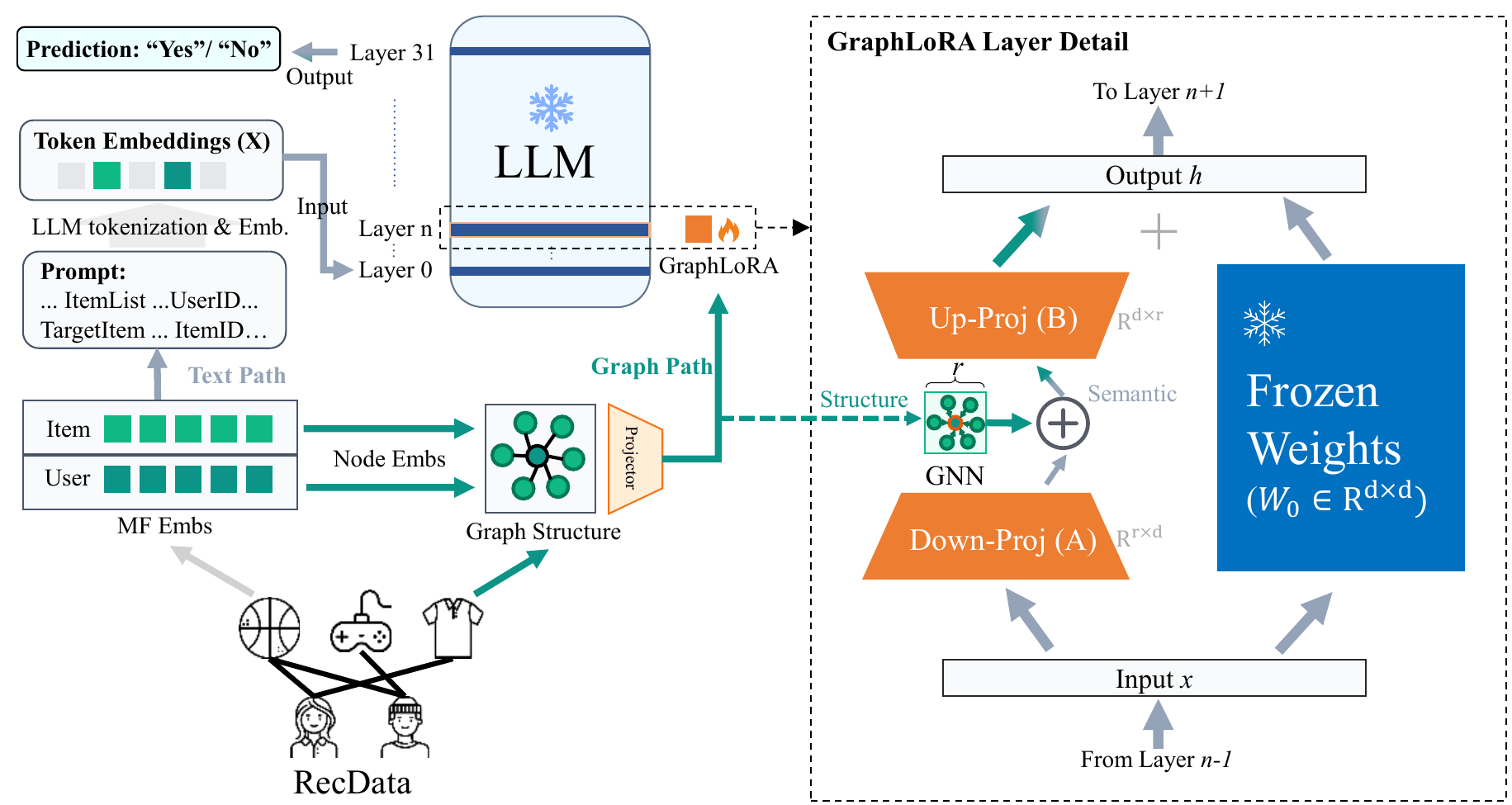}
  \caption{Overview of \textbf{GraphLoRA}. 
(Left) The end-to-end recommendation process fusing text prompts and collaborative signals. 
(Right) The detailed structure of the GraphLoRA layer, where a GNN-based structural encoder is injected between the low-rank matrices $\mathbf{A}$ and $\mathbf{B}$ to realize structure-aware low-rank propagation. 
Note: ``Embs'' refers to ``Embeddings''.}
  \label{fig:rec_model}  
\end{figure*}

\section{Methodology}

In this section, we present GraphLoRA, a unified structure-aware framework aligning collaborative representations with textual semantics.
Formally, let the input sequence be $\mathbf{X} = \{\mathbf{x}_1, \dots, \mathbf{x}_T\}$, a hybrid sequence comprising standard text tokens and collaborative tokens (i.e., users $\mathbf{x}_{userid}$ and items $\mathbf{x}_{itemid}$).
Here, $\mathbf{x}_t \in \mathbb{R}^{d_{model}}$ denotes the embedding of the $t$-th token. GraphLoRA specifically intervenes in the processing of these collaborative tokens.

As illustrated in Figure~\ref{fig:rec_model}, the workflow proceeds in four stages:
(1) \textit{Collaborative Initialization}, defining the source embeddings;
(2) \textit{Dual-View Input Construction}, mapping these embeddings into the input sequence $\mathbf{X}$;
(3) \textit{Graph Structural Encoding}, refining the embeddings via message passing; and
(4) \textit{Structure-Aware Parameter Injection}, fusing structural signals into the processing of $\mathbf{x}_t$ in the parameter space.

\subsection{Collaborative Initialization}
To ground user and item identities, we employ a set of learnable embeddings rooted in matrix factorization (MF).
We approximate the interaction matrix via latent factors $\mathbf{P} \in \mathbb{R}^{|\mathcal{U}| \times d_{emb}}$ for users and $\mathbf{Q} \in \mathbb{R}^{|\mathcal{I}| \times d_{emb}}$ for items, where $d_{emb}$ denotes the collaborative embedding dimension. The unified collaborative embedding matrix is denoted as:
\begin{equation}
    \mathbf{E} = [\mathbf{P}; \mathbf{Q}] \in \mathbb{R}^{(|\mathcal{U}| + |\mathcal{I}|) \times d_{emb}}.
\end{equation}
\textbf{Rationale.} Instead of using frozen external features, we treat $\mathbf{E}$ as a shared, learnable state. It feeds both the LLM input and the graph encoder, and is jointly optimized with the LLM. This ensures collaborative signals remain dynamically aligned with the model's semantic reasoning.

\subsection{Dual-View Input Construction}
Here, we describe how collaborative data is mapped into the LLM's input space $\mathbf{X}$.
We construct textual prompts using templates (Table~\ref{tab:prompt}). To align the collaborative dimension ($d_{emb}$) with the LLM's model dimension ($d_{model}$), we employ an input projection network $\mathcal{F}$.

For a collaborative placeholder token (e.g., \texttt{<UserID>} or \texttt{<TargetItemID>}) at position $t$ within sequence $b$, its representation $\mathbf{x}_t$ is derived directly from the current embedding state:
\begin{equation}
    \mathbf{x}_{t} = \mathcal{F}(\mathbf{e}_n), \quad \mathbf{x}_{t} \in \mathbb{R}^{d_{model}},
\end{equation}
where $\mathbf{e}_n$ corresponds to the embedding of the associated collaborative node (user or item) in $\mathbf{E}$.
We record the set of collaborative anchor positions as $\mathcal{S}=\{(b,t)\}$, where $b$ indexes the sequence (i.e., sample) within a mini-batch and $t$ denotes the token position in that sequence.
For each anchor $(b,t)\in\mathcal{S}$, we maintain a deterministic mapping $\pi(b,t)\mapsto n$ that links the collaborative token to its corresponding graph node $n$ (user or item), enabling structure injection at the correct positions.
At this stage, $\mathbf{x}_t$ contains identity information but lacks explicit topological context.

\begin{table}[t]
\centering
\renewcommand{\arraystretch}{1.2}
\begin{tabularx}{\linewidth}{X}
\toprule
\textbf{\#Question:} A user has given high ratings to the following books: \texttt{<ItemTitleList>}. Additionally, we have information about the user's preferences encoded in the feature \texttt{<UserID>}. Using all available information, make a prediction about whether the user would enjoy the book titled \texttt{<TargetItemTitle>} with the feature \texttt{<TargetItemID>}? Answer with ``Yes'' or ``No''. \\
\\[-2ex]
\textbf{\#Answer:} ``Yes'' or ``No''. \\
\bottomrule
\end{tabularx}
\caption{Example of the prompt template used in GraphLoRA.}
\label{tab:prompt}
\end{table}

\subsection{Graph Structural Encoding}
Parallel to the LLM input processing, we generate high-order structural signals.
Given the user-item subgraph and the shared embeddings $\mathbf{E}$, the graph encoder performs message passing across $L$ layers. Following the MPNN paradigm (Section~\ref{sec:mpg}), the structural representation for node $n$ evolves by aggregating messages from neighbors $\mathcal{N}_n$:
\begin{equation}
    \mathbf{e}_n^{(k+1)} = \psi \Big( \mathbf{e}_n^{(k)}, \mathop{\text{AGG}}_{v \in \mathcal{N}_n} \big( \phi(\mathbf{e}_v^{(k)}, \mathbf{e}_n^{(k)}) \big) \Big).
\end{equation}
After $L$ layers, we obtain the structure-enhanced representation $\mathbf{e}_n^{(L)} \in \mathbb{R}^{d_{emb}}$.

\textbf{Bottleneck Alignment.}
Typically, $d_{emb}$ is much larger than the LoRA intrinsic rank ($r$). To align these spaces, we employ a bottleneck projection $\mathbf{W}_{neck} \in \mathbb{R}^{r \times d_{emb}}$:
\begin{equation}
    \mathbf{z}_n = \mathbf{W}_{neck} \mathbf{e}^{(L)}_n, \quad \mathbf{z}_n \in \mathbb{R}^r,
\end{equation}
where $\mathbf{z}_n$ represents the compressed, task-specific structural signal ready for injection.

\subsection{Structure-Aware Parameter Injection}
Finally, we describe how the graph signal $\mathbf{z}_n$ is injected into the processing of $\mathbf{x}_t$. We adopt a sparse single-layer injection strategy targeting a specific layer $l^*$.

Recall that standard LoRA computes the update as $\frac{\alpha}{r}\mathbf{B}\mathbf{A}\mathbf{x}$. We intervene in the low-rank latent space between the down-projection $\mathbf{A}$ and up-projection $\mathbf{B}$.
For a collaborative token $\mathbf{x}_t$ at position $t$ (i.e., $(b,t) \in \mathcal{S}$), where the associated node is $n=\pi(b,t)$ (user or item), the injection flow is:

\noindent\textbf{1. Semantic Compression:} The input state $\mathbf{x}_t$ is compressed by matrix $\mathbf{A}$ to obtain the semantic intermediate representation $\mathbf{h}_{sem}$:
\begin{equation}
   \mathbf{h}_{sem} = \mathbf{A}\mathbf{x}_{t}, \quad \mathbf{h}_{sem} \in \mathbb{R}^r.
\end{equation}

\noindent\textbf{2. Structural Fusion:} We fuse the graph signal $\mathbf{z}_n$ with the semantic signal $\mathbf{h}_{sem}$ to produce the structure-enhanced latent state $\mathbf{h}_{latent}$:
\begin{equation}
   \mathbf{h}_{latent} = \lambda_{lora} \mathbf{h}_{sem} + \lambda_{gnn} \mathbf{z}_n,
\end{equation}
where $\lambda_{lora}, \lambda_{gnn}$ are balancing coefficients.

\noindent\textbf{3. Manifold Projection:} The fused signal is projected back to the LLM space by the up-projection matrix $\mathbf{B}$ to yield the structure-aware update, producing the final hidden state $\mathbf{h}$:
\begin{equation}
   \Delta \mathbf{x}_t = \mathbf{B} \mathbf{h}_{latent}, \quad \mathbf{h} = \mathbf{W}_0 \mathbf{x}_{t} + \frac{\alpha}{r} \Delta \mathbf{x}_t.
\end{equation}
This formulation ensures that the graph structural signal explicitly guides the gradient updates.

For non-collaborative tokens ($(b,t) \notin \mathcal{S}$), the structural pathway is inactive (i.e., $\mathbf{z}_n$ is omitted), and the model follows the standard LoRA computation.
Crucially, the GNN parameters and $\mathbf{W}_{neck}$ are jointly optimized with the LLM. During backpropagation, gradients flow through $\mathbf{B} \to \mathbf{z}_n \to \mathbf{W}_{neck} \to \text{GNN}$, ensuring that $\mathbf{z}_n$ evolves into a \textit{semantically aligned} structural bias that directly complements the reasoning over $\mathbf{x}_t$.

\subsection{Complexity Analysis}

\textbf{Parameter Efficiency.}
GraphLoRA attains structure awareness with minimal parameter overhead (e.g., $\sim 1.67\%$ over the LoRA-only baseline) via Bottleneck Alignment and Sparse Injection.
Unlike input alignment requiring the full hidden space $d_{model}$, GraphLoRA projects signals into the bottleneck $r$.
This reduces projection complexity from $\mathcal{O}(d_{model} \times d_{emb})$ to $\mathcal{O}(r \times d_{emb})$, avoiding high-dimensional redundancy.
Moreover, aligning structure with the task-relevant low-rank subspace preserves high information density and enables more effective joint optimization with LoRA parameters.
Together, these designs keep GraphLoRA lightweight while ensuring faithful and effective structural injection.

\textbf{Computational Overhead.}
The graph message passing is performed on sampled subgraphs, yielding linear complexity $O(|\mathcal{E}_{sub}|)$.
Structural embeddings are computed once per batch and injected into the target layer, decoupling structural encoding from the LLM’s depth.
As a result, GraphLoRA introduces only minimal training overhead while maintaining an effective trade-off between structural reasoning performance and computational efficiency, which is empirically validated in Section~\ref{sec:experiments}.

\begin{table}[t]
\centering
\small
\setlength{\tabcolsep}{2.5pt} 
\begin{tabular}{lccccc}
\toprule
\textbf{Dataset} & \textbf{Train} & \textbf{Valid} & \textbf{Test} & \textbf{\#Users} & \textbf{\#Items} \\
\midrule
ML-1M       & 33,891  & 10,401 & 7,331  & 839    & 3,256 \\
Amazon-Book & 727,468 & 25,747 & 25,747 & 22,967 & 34,154 \\
\bottomrule
\end{tabular}
\caption{Statistics of the processed datasets.}
\label{tab:dataset}
\end{table}

\begin{table*}[t]
\centering
\small
\renewcommand{\arraystretch}{1.0} 
\setlength{\tabcolsep}{13pt}
\resizebox{\linewidth}{!}{
\begin{tabular}{llcccc}
\toprule
\multicolumn{2}{c}{\textbf{Dataset}} & \multicolumn{2}{c}{\textbf{ML-1M}} & \multicolumn{2}{c}{\textbf{Amazon-Book}} \\
\cmidrule(lr){1-2}\cmidrule(lr){3-4} \cmidrule(lr){5-6} 
\multicolumn{2}{c}{\textbf{Methods (Settings)}} & \textbf{AUC} & \textbf{UAUC} & \textbf{AUC} & \textbf{UAUC} \\
\midrule
\multirow{3}{*}{\textbf{Collab.}} 
 & MF & 0.6486 & 0.6396 & 0.7105 & 0.5543 \\
 & LightGCN & 0.5858 & 0.6512 & 0.7026 & 0.5619 \\
 & NGCF   & 0.6248 & 0.5991 & 0.7091 & 0.5411 \\
 & SASRec & 0.7005 & 0.6734 & 0.6675 & 0.5614 \\
\midrule
\multirow{3}{*}{\textbf{LLMRec}} 
 & ICL & 0.5119 & 0.5178 & 0.5180 & 0.5043 \\
 & Prompt4NR & 0.7027 & 0.6713 & 0.6527 & 0.5011 \\
 & TALLRec & 0.7044 & 0.6741 & 0.6583 & 0.4971 \\
\midrule
\multirow{5}{*}{\shortstack[l]{\textbf{Input-Space}}} 
 & PersonPrompt & 0.7014 & 0.6503 & 0.7113 & 0.5596 \\
 & CoLLM-MF \scriptsize{$(r=8, \{q,v\})$} & 0.7028 & 0.6714 & 0.8021 & 0.5782 \\
 & CoLLM-LightGCN \scriptsize{$(r=8, \{q,v\})$} & 0.7164 & 0.6842 & 0.7835 & 0.5663 \\
 & CoLLM-SAS \scriptsize{$(r=8, \{q,v\})$} & 0.7059 & 0.6531 & 0.7538 & 0.5874 \\
 & BinLLM \scriptsize{$(r=8, \{q,v\})$} & 0.7132 & 0.6815 & 0.8157 & 0.5724 \\
\midrule
\multirow{3}{*}{\textbf{Param-Space}} 
 & CoRA-MF \scriptsize{$(r=16, \{q,k,v,o\})$}& \underline{0.7361} & 0.6884 & \underline{0.8179} & \underline{0.6262} \\
 & CoRA-LightGCN \scriptsize{$(r=16, \{q,k,v,o\})$}& 0.7128 & \underline{0.6966} & 0.7886 & 0.5689 \\
 & CoRA-SAS \scriptsize{$(r=16, \{q,k,v,o\})$}& 0.7019 & 0.6517 & 0.7677 & 0.5961 \\
\midrule
\textbf{Ours} & \textbf{GraphLoRA} \scriptsize{$(r=8, \{q,v\})$}& \textbf{0.7472} & \textbf{0.7102} & \textbf{0.8205} & \textbf{0.6303} \\
\bottomrule
\end{tabular}}
\caption{Performance comparison with state-of-the-arts. The best results are highlighted in \textbf{bold}, and the second-best results are \underline{underlined}. Parameter configurations are reported for transparency. Notably, the baselines employ their optimal reported settings (e.g., $r=16, \{q,k,v,o\}$ for CoRA), whereas GraphLoRA achieves SOTA performance under a strict, parameter-efficient constraint ($r=8, \{q,v\}$).}
\label{tab:main_exper}
\end{table*}

\section{Experiments}
\label{sec:experiments}

In this section, we evaluate the effectiveness, efficiency, and generalization of GraphLoRA. 

\subsection{Experimental Setup}
\label{sec:exp_setup}

\paragraph{Datasets.}
We conduct experiments on two standard benchmarks widely used in recommendation research: \textbf{ML-1M}\footnote{\url{https://grouplens.org/datasets/movielens/1m/}}~\cite{harper2015movielens} and \textbf{Amazon-Book}\footnote{\url{https://jmcauley.ucsd.edu/data/amazon/index_2014.html}}~\cite{he2016ups}.
To ensure a fair comparison with baseline methods, we strictly follow the data preprocessing protocols established in CoRA~\cite{CoRA}.
The detailed statistics of the processed datasets are listed in Table~\ref{tab:dataset}.

\paragraph{Evaluation Metrics.}
To comprehensively assess the recommendation performance, we adopt two widely recognized metrics: \textbf{AUC} (Area Under the ROC Curve) and \textbf{UAUC} (User-averaged AUC)~\cite{liu2021concept}.
AUC evaluates the global ranking capability of the model across all samples, while UAUC calculates the AUC for each user individually and reports the average.

\paragraph{Implementation Details.}
We employ Vicuna-7B~\cite{vicuna2023} as the backbone LLM for our main experiments.
The collaborative encoder uses MF ($d=256$) and a 3-layer NGCF~\cite{wang2019neural} to capture structural signals. 
To balance high-fidelity collaborative signals with extreme parameter efficiency, we adopt a first-order (1-hop) neighbor sampling strategy for direct GNN aggregation.
We set the LoRA rank to $r=8$ and adapt the query ($q$) and value ($v$) projections.
GraphLoRA adopts a sparse single-layer injection strategy.
Based on validation tuning, we inject structural signals into the 31st layer for ML-1M and the 15th layer for Amazon-Book, with $\lambda_{lora}=1.0$ and $\lambda_{gnn}=0.1$.

All experiments are conducted on 4 NVIDIA RTX 3090 (24GB) GPUs.
We train end-to-end with BCE loss, using AdamW (for the LLM) and Adam (for the GNN).
We grid-search learning rates in $\{5e^{-4}, 1e^{-4}, 5e^{-5}\}$ and weight decay in $\{1e^{-3}, 1e^{-4}\}$.
Unless otherwise specified, we use a global batch size of 8, learning rate $1e^{-4}$, cosine scheduling with warm-up, and maximum sequence length 1024.

\paragraph{Baselines.}
We compare GraphLoRA with baselines from four paradigms:
\textbf{(1) Conventional Collaborative Filtering:} MF~\cite{mf}, NGCF~\cite{wang2019neural}, LightGCN~\cite{he2020lightgcn}, and SASRec~\cite{kang2018self}.
\textbf{(2) LLM-based Recommendation:} ICL~\cite{dai2023uncovering}, Prompt4NR~\cite{zhang2023prompt}, and TALLRec~\cite{TALLRec}.
\textbf{(3) Input-Space Alignment:} PersonPrompt~\cite{li2023personalized}, CoLLM~\cite{Collm2025}, and BinLLM~\cite{zhang2024binllm}.
\textbf{(4) Parameter-Space Alignment:} CoRA~\cite{CoRA}, including CoRA-MF, CoRA-LightGCN, and CoRA-SAS.

\textbf{Baseline configurations.}
We follow the default settings in the original papers and/or official implementations, and tune the remaining hyper-parameters based on validation AUC.
Adapter configurations are kept as in the original methods (e.g., CoRA uses $r=16$ on $\{q,k,v,o\}$, while CoLLM/TALLRec/BinLLM use $r{=}8$ on $\{q,v\}$), and we report them in tables for transparency.

\subsection{Performance Comparison}
\label{sec:overall_perf}
\begin{figure}[t]
  \centering
  \includegraphics[width=\linewidth]{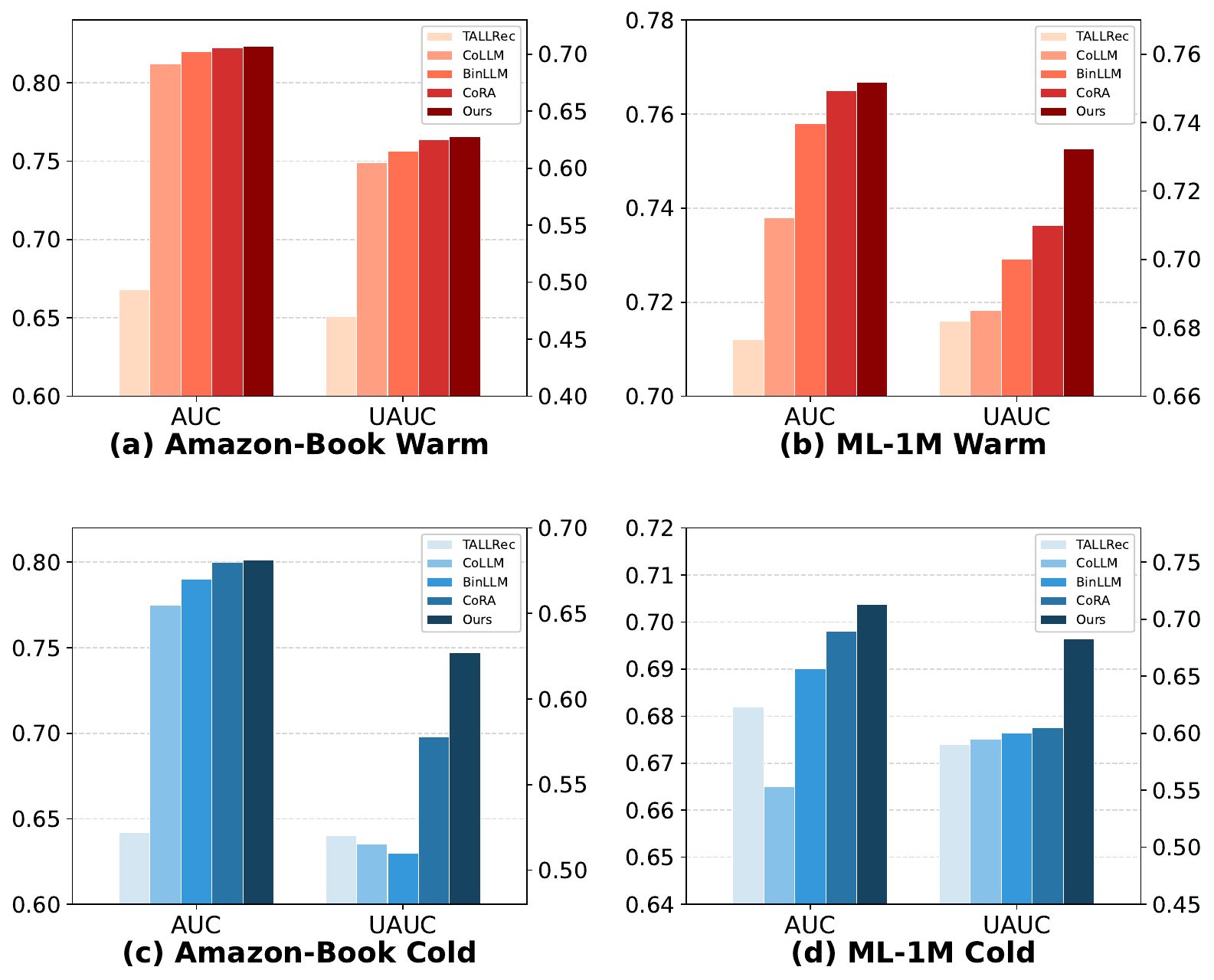}
  \caption{Warm/cold evaluation on Amazon-Book and ML-1M. The left and right y-axes correspond to AUC and UAUC, respectively.}
  \label{fig:warm_cold}
\end{figure}

We compare GraphLoRA with representative baselines on ML-1M and Amazon-Book.
Table~\ref{tab:main_exper} reports overall results, and Figure~\ref{fig:warm_cold} further decomposes performance under warm/cold start regimes.

\paragraph{Overall Results.}
GraphLoRA consistently achieves the best performance across both datasets and all evaluation metrics.
Compared with strong \textit{input-space alignment} baselines, it delivers substantial gains, indicating that injecting collaborative signals directly into the parameter space enables tighter integration between structural information and prompt-conditioned semantics than shallow input augmentation.

\paragraph{Comparison with Parameter-Space Alignment.}
We compare GraphLoRA with the parameter-space alignment baseline CoRA.
While CoRA injects high-capacity adaptations across all Transformer layers, GraphLoRA uses a sparse, targeted injection with a small rank(e.g., $r{=}8$) at a single-layer.
Despite this compact design, GraphLoRA achieves comparable or better performance, highlighting a more favorable performance–efficiency trade-off.

\paragraph{Warm and Cold Start Analysis.}
We split the test set into warm and cold subsets based on user interaction frequency (Figure~\ref{fig:warm_cold}).
In warm-start, GraphLoRA achieves performance comparable to the strongest baseline, whereas in the cold-start scenario, it exhibits a more pronounced advantage.
This observation aligns with our design intuition: the jointly optimized graph encoder aggregates neighborhood context, providing informative structural signals that help alleviate data sparsity for users with limited interaction histories.

\paragraph{Matched-Budget Fairness and Top-K Ranking.}
To further validate that GraphLoRA's superiority stems from its structure-aware injection mechanism rather than parameter configuration advantages, we re-evaluate the strongest baseline, CoRA-MF, under the exact same constrained parameter budget as GraphLoRA ($r=8, \{q,v\}$). Furthermore, to provide a comprehensive evaluation beyond binary prediction, we introduce NDCG@10 to assess top-K ranking performance. 

As shown in Table~\ref{tab:matched_budget}, when the parameter budget is strictly restricted, the baseline CoRA-MF experiences a severe "perception collapse" (e.g., UAUC dropping to 0.4995 on Amazon-Book). In contrast, GraphLoRA remains highly effective under identical low-rank constraints. Moreover, GraphLoRA consistently achieves superior NDCG@10 scores, demonstrating its ability to robustly internalize collaborative topology and translate it into high-quality top-list recommendations without relying on parameter inflation.

\begin{table}[htb]
\centering
\small
\setlength{\tabcolsep}{3pt}
\resizebox{\columnwidth}{!}{
\begin{tabular}{llcc}
\toprule
\textbf{Dataset} & \textbf{Metric} & \textbf{\shortstack{CoRA-MF\\$(r=8, \{q,v\})$}} & \textbf{\shortstack{GraphLoRA (Ours)\\$(r=8, \{q,v\})$}} \\
\midrule
\multirow{3}{*}{\textbf{ML-1M}} 
 & AUC & 0.7227 & \textbf{0.7472} \\
 & UAUC & 0.6794 & \textbf{0.7102} \\
 & NDCG@10 & 0.7636 & \textbf{0.7847} \\
\midrule
\multirow{3}{*}{\textbf{Amazon-Book}} 
 & AUC & 0.7562 & \textbf{0.8205} \\
 & UAUC & 0.4995 & \textbf{0.6303} \\
 & NDCG@10 & 0.6990 & \textbf{0.7067} \\
\bottomrule
\end{tabular}
}
\caption{Matched-budget comparison and Top-K ranking performance. Both methods are evaluated under identical parameter constraints ($r=8$ on $\{q,v\}$).}
\label{tab:matched_budget}
\end{table}

\subsection{Ablation and Efficiency Analysis}
\label{sec:ablation}

\begin{table}[t]
\centering
\resizebox{\columnwidth}{!}{
    \setlength{\tabcolsep}{1.2mm}
    \begin{tabular}{lcccc}
    \toprule 
    \multirow{2}{*}{\textbf{Method Settings}} &
    \multicolumn{2}{c}{\textbf{ML-1M}} &
    \multicolumn{2}{c}{\textbf{Amazon-Book}} \\
    \cmidrule(lr){2-3} \cmidrule(lr){4-5}
    & \textbf{AUC} & \textbf{UAUC} & \textbf{AUC} & \textbf{UAUC} \\
    \midrule
    \multicolumn{5}{l}{\textit{\textbf{(A) Training Strategy \& Components}}} \\
    \addlinespace[1.2ex]
    1. LoRA-only(Frozen MF)       & 0.6981 & 0.6548 & 0.8012 & 0.6026 \\
    2. LoRA-only(Trainable MF) & 0.7178 & 0.6952 & 0.8136 & 0.6153 \\
    \textbf{3. GraphLoRA (Ours)} & \textbf{0.7472} & \textbf{0.7102} & \textbf{0.8205} & \textbf{0.6303} \\
    \midrule
    \multicolumn{5}{l}{\textit{\textbf{(B) Graph Encoder Variants}}} \\
    \addlinespace[1.2ex]
    GraphLoRA (GCN)      & 0.7417 & 0.6934 & 0.8168 & 0.6277 \\
    GraphLoRA (LightGCN) & 0.7382 & 0.7034 & 0.8132 & 0.6170 \\
    GraphLoRA (NGCF)     & \textbf{0.7472} & \textbf{0.7102} & \textbf{0.8205} & \textbf{0.6303} \\
    \bottomrule
    \end{tabular}
}
\caption{Ablation of (A) training strategy/components and (B) graph encoder variants on ML-1M and Amazon-Book (AUC/UAUC).}
\label{tab:ablation}
\end{table}

\paragraph{Training Strategy and Components.}
We examine progressive settings motivated by the two roles of MF in GraphLoRA: it anchors user/item identities in the prompt (via input projection) and provides initial states for graph aggregation.
Specifically, (1) \textit{LoRA-only (Frozen MF)} freezes MF embeddings and disables the graph-injection path;
(2) \textit{LoRA-only (Trainable MF)} jointly optimizes MF with the LLM, and uses the injection path \emph{without} graph aggregation (identity mapping);
(3) \textit{GraphLoRA} is the full model with a graph encoder integrated into the LoRA bottleneck.
As shown in Table~\ref{tab:ablation}(A), freezing MF yields the weakest performance.
Making MF trainable improves the results, indicating that updating collaborative representations under the recommendation objective is beneficial.
Building on this stronger starting point, the full GraphLoRA further boosts performance by enabling topology-aware aggregation and injecting the refined structural signal inside the LoRA bottleneck.

\paragraph{Graph Encoder Variants.}
We vary the graph encoder inside the same injection mechanism (Table~\ref{tab:ablation}(B)).
Among GCN, LightGCN, and NGCF, NGCF achieves the best results in our setting, suggesting that richer interaction modeling can be beneficial for producing the injected structural signal.

\begin{table}[t]
\centering
\footnotesize
\small
\setlength{\tabcolsep}{1.2mm}
\renewcommand{\arraystretch}{1.00}
\begin{tabularx}{\columnwidth}{>{\raggedright\arraybackslash}Xccccc}
\toprule
\textbf{Method} & \textbf{Params} & \textbf{Tr.} & \textbf{Inf.} & \textbf{AUC} & \textbf{UAUC} \\
\midrule
\multicolumn{6}{l}{\textit{\textbf{(A) Baselines \& Overall}}} \\
\addlinespace[0.6ex]
\shortstack[l]{LoRA-only\\\scriptsize$(r=8,\{q,v\})$}
& 1.000 & 1.000 & 1.000 & 0.7178 & 0.6952 \\

\shortstack[l]{CoRA-MF\\\scriptsize$(r=16,\{q,k,v,o\})$\hspace{0.4em}}
& 7.635 & 1.086 & 1.009 & 0.7361 & 0.6884 \\

\shortstack[l]{GraphLoRA (Ours)\\\scriptsize$(r=8,\{q,v\})$}
& \textbf{1.017} & \textbf{1.029} & \textbf{1.014} & \textbf{0.7472} & \textbf{0.7102} \\

\midrule
\multicolumn{6}{l}{\textit{\textbf{(B) Injection Position Ablation}}} \\
\addlinespace[0.6ex]
\shortstack[l]{Pre-$A$\\\scriptsize(GNN before $A$)}
& 5.224 & 1.058 & 1.033 & 0.7333 & 0.6941 \\
\textbf{\shortstack[l]{Middle (Ours)\\\scriptsize($A \rightarrow$ GNN $\rightarrow B$)}}
& \textbf{1.017} & \textbf{1.029} & \textbf{1.014} & \textbf{0.7472} & \textbf{0.7102} \\

\bottomrule
\end{tabularx}
\caption{Efficiency and injection-position analysis on ML-1M. Params/Tr./Inf. are overhead ratios (Base = 1.00) for trainable parameters, training time, and inference time, respectively.}
\label{tab:efficiency_relative}
\end{table}

\paragraph{Efficiency and Injection Position.}
Table~\ref{tab:efficiency_relative} reports overheads normalized by the LoRA-only baseline (with trainable MF).
GraphLoRA achieves superior performance with negligible costs ($1.017\times$ parameters, $\sim1\%$ latency), significantly outperforming the heavy-budget CoRA-MF ($7.635\times$).
Regarding injection position, the \emph{Pre-$A$} strategy operates in the high-dimensional space ($d_{model}$), causing parameter explosion ($5.224\times$).
In contrast, our \textbf{Middle} strategy targets the low-rank bottleneck ($r \ll d_{model}$).
This design offers a \textbf{highly efficient structure-aware} alternative: by fusing signals within the task-specific bottleneck, it enables the graph topology to guide semantic adaptation compactly, fostering deep structure--semantic synergy within a constrained manifold.

\begin{figure}[t]
  \centering
  \includegraphics[width=0.9\linewidth]{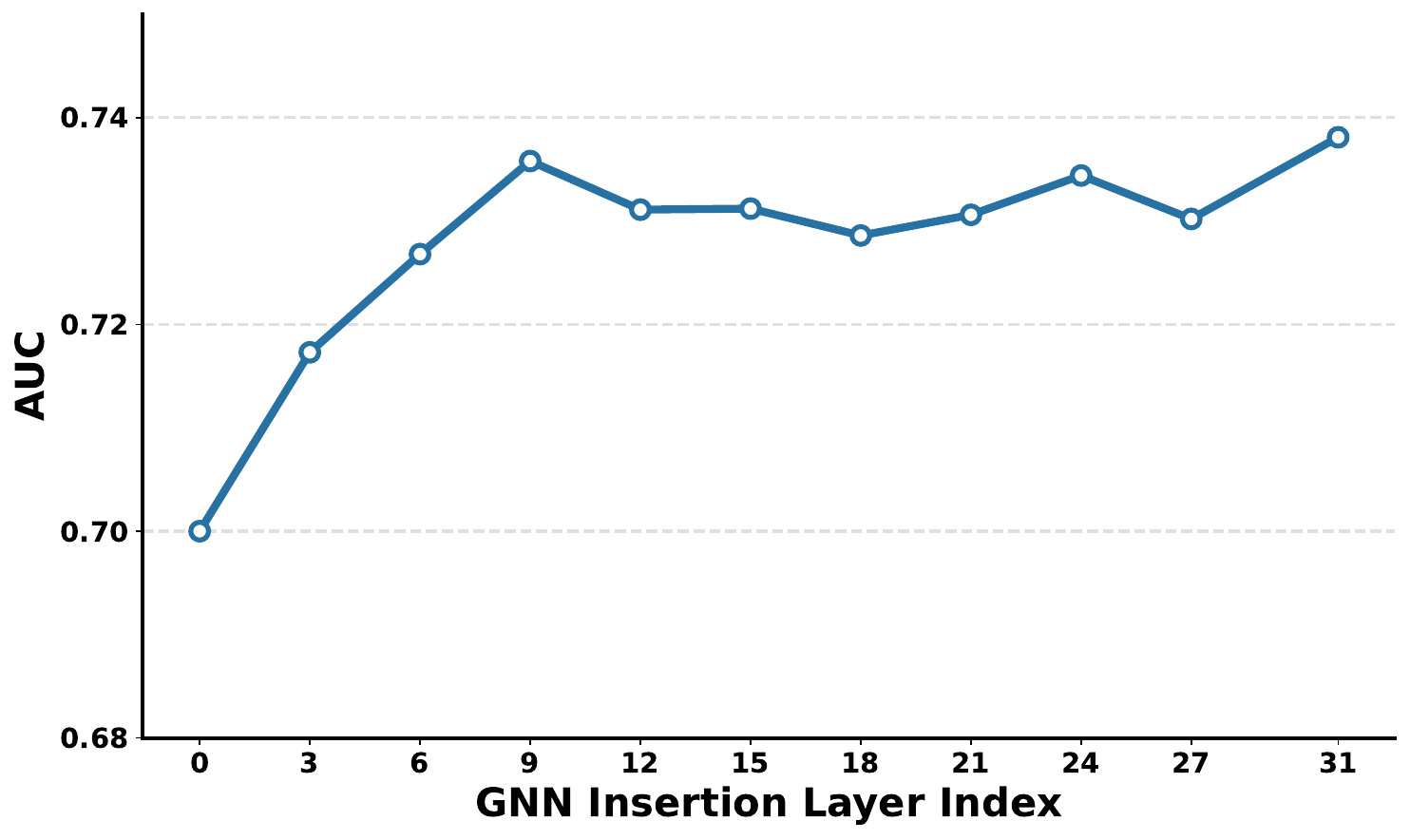}
  \caption{Effect of the GNN insertion layer on ML-1M (validation AUC).}
  \label{fig:auc_layer}
\end{figure}

\paragraph{Injection Layer Depth.}
We study how injection depth affects GraphLoRA by inserting the GNN module at different Transformer layers.
On ML-1M, we sweep the insertion layer index and report validation AUC (Figure~\ref{fig:auc_layer}).
Deeper-layer injection is generally more effective, with the best performance observed near the top layers (e.g., layer 31 for ML-1M and layer 15 for Amazon-Book). This is consistent with recent interpretability research~\cite{jin-etal-2025-exploring,skean2024does}, which demonstrates that LLMs process information hierarchically: shallower layers primarily handle surface-level syntax, while deeper intermediate layers encode more abstract, task-relevant semantics optimal for aligning topology-aware collaborative signals.

\subsection{Generalization Analysis}
\label{sec:generalization}

\paragraph{Few-Shot Performance and Backbone Robustness.}
To verify GraphLoRA robustness and generalization across architectures, we adopt the few-shot protocol from TALLRec~\cite{TALLRec} and switch the backbone from Vicuna to \textbf{LLaMA-7B}.
Experiments are conducted on the Book-Crossing~\cite{BookCrossing} dataset with 16, 64, and 256 training samples.
As shown in Table~\ref{tab:few_shot}, GraphLoRA consistently outperforms baselines across all regimes.
This confirms that our topology-aware injection is not backbone-specific and effectively complements LLM reasoning even when explicit supervision is scarce.

\begin{table}[t]
\centering
\small
\setlength{\tabcolsep}{2pt}
\begin{tabular}{ccccc}
\toprule
\textbf{Few-shot} & \textbf{SASRec} & \textbf{GRU-BERT} & \textbf{TALLRec} & \textbf{GraphLoRA} \\
\midrule
16   & 49.48 & 50.07 & 56.36 & \textbf{60.90} \\
64   & 50.06 & 49.64 & 60.39 & \textbf{61.08} \\
256  & 50.20 & 49.79 & 64.38 & \textbf{67.90} \\
\bottomrule
\end{tabular}
\caption{Few-shot AUC on Book-Crossing under the TALLRec evaluation protocol }
\label{tab:few_shot}
\end{table}

\section{Conclusion}
In this study, we present GraphLoRA, a structure-aware framework that bridges collaborative signals and textual semantics within LLM.
By embedding a graph message-passing module into the LoRA adaptation pathway, GraphLoRA generalizes traditional low-rank adaptation to structure-aware propagation, enabling joint reasoning over graph topology and textual semantics.
Through unified optimization, it distills neighborhood-aware structural signals without disrupting linguistic competence.
Extensive experiments across multiple datasets and evaluation settings demonstrate the effectiveness and efficiency of GraphLoRA, highlighting the potential of structure-aware parameter adaptation for future LLM-based recommendation research.

\section{Limitations}
This work provides an initial step towards structure-aware parameter-efficient adaptation for LLM-based recommendation.
There remain several directions that are not fully explored in this paper.
First, while we evaluate on representative benchmarks and backbones, validating GraphLoRA on a broader range of LLMs, domains, and recommendation tasks would further strengthen the generality of our conclusions.
Second, our current study focuses on a targeted injection design; more systematic analysis of architectural choices (e.g., injection depth and module variants) across settings is left for future work.
Finally, incorporating additional evaluation dimensions (e.g., robustness under different data distributions or efficiency under larger-scale deployments) would provide a more comprehensive understanding of the method's behavior.

\section{Ethics Statement}

\textbf{Data Privacy.} We utilize public, anonymized benchmarks devoid of Personally Identifiable Information (PII), strictly complying with their respective licenses. No private user data was collected or processed.

\textbf{Biases and Societal Impact.} While our work focuses on structural alignment, LLM-based recommenders inherently risk amplifying biases or creating echo chambers. Real-world deployments require rigorous fairness and safety evaluations to mitigate potential discrimination.

\textbf{Environmental Impact.} GraphLoRA promotes ``Green AI.'' By freezing the LLM backbone and tuning only a minimal parameter set, our approach significantly reduces energy consumption and carbon footprint compared to full-model fine-tuning, supporting sustainable research.

\section{Acknowledgements}
This work is supported by the National Natural Science Foundation of China (No.62206004, No.62572002, No.62272001, No.624065095), and the Natural Science Foundation of Anhui Province (No.2208085QF199, No.2508085MF159, No.2308085MF213).

\bibliography{acl}
\end{document}